\documentclass{article} 
\usepackage{arxiv,times}

\usepackage{amsmath,amsfonts,bm}

\def\eqref#1{equation~\ref{#1}}

\def\1{\bm{1}}

\DeclareMathAlphabet{\mathsfit}{\encodingdefault}{\sfdefault}{m}{sl}
\SetMathAlphabet{\mathsfit}{bold}{\encodingdefault}{\sfdefault}{bx}{n}

\DeclareMathOperator*{\argmax}{arg\,max}

\usepackage{mathtools}
\usepackage{bbm}
\usepackage{bm}
\usepackage{dsfont}

\usepackage{tablefootnote}

\usepackage{xparse}
\ExplSyntaxOn

\NewDocumentCommand{\definealphabet}{mmmm}
 {
  \int_step_inline:nnn { `#3 } { `#4 }
   {
    \cs_new_protected:cpx { #1 \char_generate:nn { ##1 }{ 11 } }
     {
      \exp_not:N #2 { \char_generate:nn { ##1 } { 11 } }
     }
   }
 }

\ExplSyntaxOff

\definealphabet{bb}{\mathbb}{A}{Z}
\definealphabet{cal}{\mathcal}{A}{Z}
\definealphabet{bf}{\bm}{a}{z}
\definealphabet{bf}{\bm}{A}{Z}

\newcommand{\inbraces}[1]{\left\{ #1 \right\}}
\newcommand{\set}[1]{\inbraces{#1}}
\newcommand{\norm}[1]{\left\lVert #1 \right\rVert}

\usepackage{hyperref}
\usepackage{url}
\usepackage{cleveref}

\usepackage{booktabs}
\usepackage{multirow}
\usepackage{geometry}

\usepackage{array}
\usepackage{adjustbox} 
\usepackage{graphicx}
\usepackage{xspace}

\newcommand{\aiscivision}{AISciVision\xspace}
\newcommand{\visrag}{VisRAG\xspace}
\newcommand{\dsaquaculture}{Aquaculture\xspace}
\newcommand{\dseelgrass}{Eelgrass\xspace}
\newcommand{\dssolar}{Solar\xspace}

\newcommand{\xtest}{\bfx_{\text{test}}}
\newcommand{\etest}{\bfe_{\text{test}}}
\newcommand{\bfepos}{\bfe_{\text{sim}}^+}
\newcommand{\bfeneg}{\bfe_{\text{sim}}^-}
\newcommand{\bfxpos}{\bfx_{\text{sim}}^+}
\newcommand{\bfxneg}{\bfx_{\text{sim}}^-}

\title{\aiscivision: A Framework for Specializing Large Multimodal Models in Scientific Image Classification}

\author{%
Brendan Hogan$^{1}$, Anmol Kabra$^{1}$, Felipe Siqueira Pacheco$^{2}$, Laura Greenstreet$^{1}$, Joshua Fan$^{1}$, \And
Aaron Ferber$^{1}$, Marta Ummus$^{3}$, Alecsander Brito$^{3}$, Olivia Graham$^{2}$, Lillian Aoki$^{4}$, Drew Harvell$^{2}$, \And
Alex Flecker$^{2}$, Carla Gomes$^{1}$
\\
\\
$^{1}$Department of Computer Science, Cornell University, Ithaca, NY 14850, USA \\
$^{2}$Department of Ecology and Evolutionary Biology, Cornell University, Ithaca, NY 14850, USA \\
$^{3}$Empresa Brasileira de Pesquisa Agropecuária, Brasília, DF, Brazil \\
$^{4}$Department of Environmental Studies, University of Oregon, Eugene, OR 97403, USA \\
\\
\texttt{bhr54@cornell.edu, anmol@cs.cornell.edu, felipe.pacheco@cornell.edu} \\
\texttt{\{leg86,jy6,amf272\}@cornell.edu, \{marta.ummus, alecsander.brito\}@embrapa.br} \\
\texttt{ojg5@cornell.edu, laoki@uoregon.edu, \{cdh5,asf3,cpg5\}@cornell.edu}
}

\finalcopy 
\usepackage{pdfpages}

\begin{document}

\maketitle

\begin{abstract}
Trust and interpretability are crucial for the use of Artificial Intelligence (AI) in scientific research, but current models often operate as black boxes offering limited transparency and justifications for their outputs. Motivated by this problem, we introduce \aiscivision, a framework that specializes Large Multimodal Models (LMMs) into interactive research partners and classification models for image classification tasks in niche scientific domains. Our framework uses two key components: (1) Visual Retrieval-Augmented Generation (VisRAG) and (2) domain-specific tools utilized in an agentic workflow. To classify a target image, \aiscivision first retrieves the most similar positive and negative labeled images as context for the LMM. Then the LMM agent actively selects and applies tools to manipulate and inspect the target image over multiple rounds, refining its analysis before making a final prediction. These VisRAG and tooling components are designed to mirror the processes of domain experts, as humans often compare new data to similar examples and use specialized tools to manipulate and inspect images before arriving at a conclusion. Each inference produces both a prediction and a natural language transcript detailing the reasoning and tool usage that led to the prediction. We evaluate \aiscivision on three real-world scientific image classification datasets: detecting the presence of aquaculture ponds, diseased eelgrass, and solar panels. Across these datasets, our method outperforms fully supervised models in low and full-labeled data settings. \aiscivision is actively deployed in real-world use, specifically for aquaculture research, through a dedicated web application that displays and allows the expert users to converse with the transcripts. This work represents a crucial step toward AI systems that are both interpretable and effective, advancing their use in scientific research and scientific discovery.
Our code is available at \url{https://github.com/gomes-lab/AiSciVision}.
\end{abstract}

\section{Introduction}
Until recently, meaningful interactions with AI models were largely restricted to researchers and practitioners, who accessed these models through technical interfaces, often for niche applications. But the emergence of Large Multimodal Models (LMMs) such as OpenAI's GPT~\citep{achiam2023gpt}, Google's Gemini~\citep{geminiteam2024geminifamilyhighlycapable}, and Meta's Llama~\citep{touvron2023llamaopenefficientfoundation} has dramatically transformed the landscape. Now, both experts and the general public can converse meaningfully with AI, making these interactions a part of everyday life. This change highlights not only the rapid advancements in model capabilities but also introduces new standards for how we think about and interact with AI, as accessible, versatile, and personal assistants.

However, while this transformation has enabled AI to serve as a general-purpose assistant across a wide range of topics~\citep{kiros2014multimodal,vinyals2015show,ramesh2021zero,abdelhamed2024you}, it raises a critical question: can these models provide the same depth of expertise in highly specialized and impactful domains? In areas such as medicine, law, scientific research and scientific discovery, the need goes beyond general conversations, these fields demand models capable of deep, domain-specific reasoning~\citep{lu2022learn,mall2024remote}. The general knowledge embedded in LMMs falls short of the nuanced expertise required for these specialized tasks, limiting their effectiveness where it matters most.

\begin{figure*}[!t]
    \centering
    \includegraphics[width=.7\textwidth]{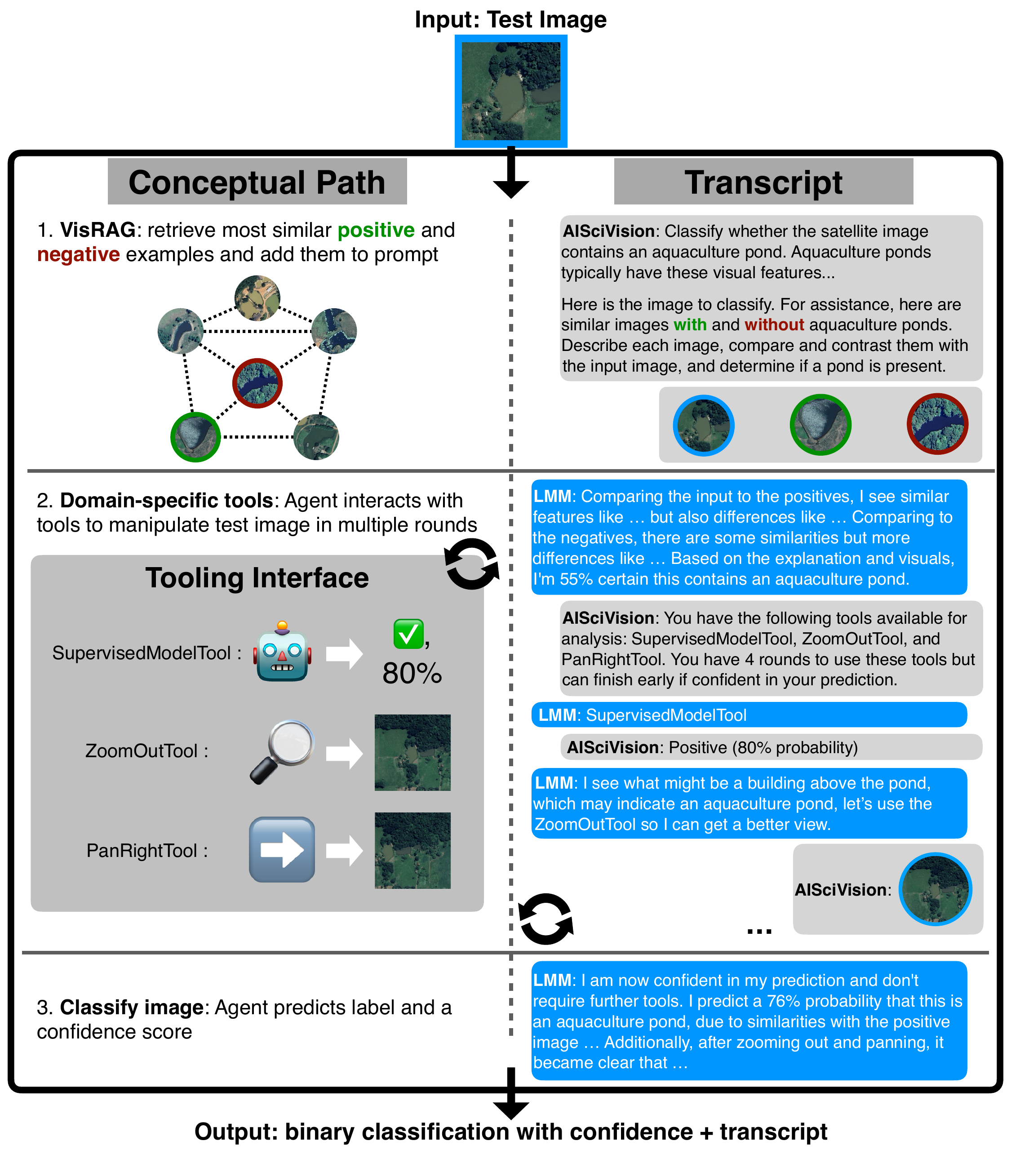}
    \caption{
    A schematic of our \textbf{\aiscivision} framework, comprising of two components (1) Visual Retrieval-Augmented Generation (\visrag) and (2) Domain-specific Tools.
    Given an input test image, we use the \visrag component to retrieve the most similar positive and negative examples from the training dataset. We then prompt the LMM to compare these images with the test image, and utilize domain-specific interactive tools over several rounds in a conversation where the \textit{LMM acts as an AI agent}.
    Finally, the agent outputs a prediction and the inference transcript.
    The transcript offers insight into the agent's reasoning process, improving interpretability, transparency, and trust, all crucial for applications in scientific domains.
    }
    \label{fig:schematic}
\end{figure*}

Fortunately, the large context windows of LMMs allow for flexible specialization via in-context learning. By providing rich prompts and context relevant to a particular task, LMMs can adapt to domain-specific requirements, a strategy that is driving exciting research in Retrieval-Augmented Generation (RAG)~\citep{lewis2020rag,khandelwal2020generalization}. RAG techniques enhance LMM predictions by retrieving task-specific examples, effectively refining the model's responses based on context and thereby specializing it for the task at hand. This approach is particularly valuable because the sheer scale and cost of training LMMs are often only feasible for large organizations. For most researchers, leveraging context is a practical way to harness the general knowledge of LMMs while also enabling them to excel at unique, specialized tasks~\citep{soong2024improving,zakka2024almanac}. One critical area for specialization is scientific image classification. The ability to adapt LMMs to these domains would be transformative, as it has the potential to democratize access to advanced analysis previously limited by the scarcity of domain expertise. By enabling AI to provide nuanced, explainable insights in specialized scientific tasks, we can significantly accelerate research and discovery, bridging the gap between powerful, general-purpose AI models and the pressing, domain-specific challenges faced by experts.

To address these challenges, we introduce \textbf{\aiscivision}, a framework that adapts general-purpose LMMs to accurately classify scientific images while generating transparent, context-specific reasoning transcripts. \aiscivision operates in a model-agnostic way, combining two key components: a Visual Retrieval-Augmented Generation system (VisRAG) and an interactive tooling AI agent that classifies images through dialogue. Users start by providing labeled training data, which is embedded into a feature space, where the system organizes positive and negative class examples separately. They also specify classification tools, ranging from basic image adjustments (e.g., contrast) to domain-specific operations like zooming into satellite imagery. During inference, AISciVision retrieves similar positive and negative image examples from the training set based on cosine similarity in the embedding space and uses them as context for the LMM’s analysis. The LMM then engages in multiple rounds of interactive analysis, using the specified tools to refine its understanding of the target image, before arriving at a final prediction. A visual schematic of AISciVision is provided in \Cref{fig:schematic}.

We evaluate our method on three real-world scientific image classification datasets: detecting aquaculture ponds in satellite images~\citep{greenstreet2023detecting}, diseased eelgrass ~\citep{rappazzo2021eelisa}, and solar panels in satellite images~\citep{lacoste2023geobenchfoundationmodelsearth}. Our \aiscivision framework outperforms both fully supervised and zero-shot methods while producing transcripts detailing the agent's reasoning. We deploy our framework as a web application for real-time scientific monitoring, where users can interact with the inference transcripts in a `Chat-GPT' style and can ask clarifying questions and provide corrections/feedback. Future work will study how this feedback can be incorporated into the VisRAG, to allow for the model to ever-improve and learn as experts converse with it. Our contributions are:
\begin{enumerate}
    \item We introduce \textbf{\aiscivision}, a novel framework for specializing Large Multimodal Models (LMMs) to niche scientific image classification tasks. It combines Visual Retrieval-Augmented Generation (\visrag) with domain-specific interactive tools, designed to emulate how human experts manipulate and inspect images. Through multi-round conversations using these tools, the LMM acts as an AI agent. To our knowledge, \aiscivision is among the first to apply such techniques to specialized scientific applications.

    \item We demonstrate the framework's efficacy on three real-world scientific image classification datasets, highlighting that our framework is flexible and easily extendable to new applications.
    We outperform several fully supervised models and zero-shot approaches, while additionally providing inference transcripts.
    These transcripts enhance interpretability, transparency, and trust in model predictions by making the entire reasoning process accessible.

    \item We have deployed \aiscivision as a web application that ecologists and scientists use to classify images and generate inference transcripts.
    The web application lays the groundwork for collecting rich and nuanced expert feedback about the LMM agent's reasoning process, opening up potential future work to incorporate such feedback to improve the agent's performance.

\end{enumerate}

\section{Related Works}

Our work builds on recent research in multimodal models, Retrieval-Augmented Generation, and interactive AI agents that leverage tools.
We discuss and compare related work in this section, and distinguish our work as integrating these domains into a unified approach.

\paragraph{Multimodal models in low-labeled data regimes}
Large Multimodal Models (LMMs) like CLIP~\citep{radford2021learning}, GPT-4~\citep{achiam2023gpt}, LLaVa~\citep{liu2023llava}, and PaLM-E~\citep{driess2023palme} have been demonstrated to understand and generate content across multiple modalities like text, visual, and audio.
By building rich and general-purpose representations of inputs from large and diverse datasets, these LMMs demonstrate competitive zero- and few-shot capabilities on a wide range of tasks~\citep{brown2020language}, such as natural language understanding~\citep{kiros2014multimodal}, image captioning~\citep{vinyals2015show}, text-to-image generation~\citep{ramesh2021zero}, and image classification~\citep{guillaumin2010multimodal,abdelhamed2024you}.
\citet{acosta2022multimodal,moor2023foundation,wang2023ScientificDI} demonstrate that these capabilities are key for advancing research in domains where obtaining labeled data is costly and tedious, for instance, in scientific research.
On the other hand, \citet{lu2022learn,mall2024remote} find that zero- and few-shot capabilities of general-purpose LMMs like CLIP do not suffice in scientific applications, where utilizing domain-specific information becomes important.
\citet{lu2022learn} find that Chain-of-Thought prompting~\citep{wei2022chain} improves question-answering performance, whereas \citet{mall2024remote} demonstrate that a ``Vision Language Model'' explicitly trained on satellite images outperforms CLIP. 
Our framework \aiscivision extends general-purpose LMMs to classify images in low-labeled data regimes like scientific applications.
We incorporate Retrieval-Augmented Generation and domain-specific tool use to ground outputs in such specialized scientific applications.
\aiscivision allows an LMM to predict after multiple rounds of tool use, thus going beyond classical Chain-of-Thought prompting.

\paragraph{Retrieval-Augmented Generation (RAG)}
Since Large Language Models (LLMs) suffer from hallucinations in generations and static memory about the world, there is much work in Retrieval-Augmented Genereation (RAG) to retrieve relevant context from external knowledge sources to ground generations in reality~\citep{khandelwal2020generalization,lewis2020rag}.
RAG has proven useful for language generation and question answering in scientific applications such as biomedical research and medicine~\citep{soong2024improving,zakka2024almanac,bae2024ehrxqa}.
Recent work has also demonstrated the effectiveness of RAG in multimodal settings by retrieving relevant images and documents to enrich the model's prompt, in applications of visual question answering and image captioning~\citep{chen2022murag,yasunaga2023rag_multimodal,lin2023fine}.
Our framework \aiscivision leverages RAG for scientific image classification, by enriching the LMM's context with domain-relevant examples during inference.

\paragraph{Interactive AI Agents and tool-use}
In recent years, Large Language Models have enabled users to engage in multi-turn natural language conversations to perform a wide range of tasks, from brainstorming and writing to writing code and solving math equations~\citep{achiam2023gpt,driess2023palme,touvron2023llamaopenefficientfoundation,geminiteam2024geminifamilyhighlycapable}.
There is growing interest in enabling such models to \textit{act as an agent} on their generations, by deploying them in environments~\citep{fan2022minedojo,wang2024voyager} and attaching tools to interact with the web~\citep{nakano2021webgpt,patil2023gorilla,schick2023toolformer}.
\citet{yao2023react} accomplish this with ReAct, by prompting the model to invoke calls to specified tools in natural language format, whereas \citet{schick2023toolformer} introduce ToolFormer which is a model finetuned on a list of available tools.
When environments do not return natural language feedback, e.g. when the feedback is scalar or binary, \citet{shinn2023reflexion} find that training a helper model to generate natural language descriptions of the feedback improves the agent's performance---termed as Reflexion.
\citet{lu2024aiscientist} introduce the ``AI Scientist'' that imitates the research process in the Machine Learning community.
They utilize Chain-of-Thought, Reflexion, and tool use to imitate the research process: brainstorm research ideas, execute experiments, and write a paper.

For scientific research in physical, life, and climate sciences, not only is it important to obtain accurate predictions but also to get insight into the underlying justifications that lead to these predictions~\citep{wang2023ScientificDI,chen2021automating,royalsociety2024science,kong2022density}.
With the ReAct approach of interacting with external tools and knowledge sources in a multi-turn conversation, \citet{yao2023react} shows that the agent leaves a `paper trail' or a `transcript' of the decision-making process.
Our framework \textbf{\aiscivision accomplishes exactly this feat, and is one of the first attempts at developing an interactive AI agent for scientific image classification}.
Following the extensive research in tool use, we develop domain-specific tools in the \aiscivision framework, for instance, zoom/pan for satellite image datasets and image enhancement tools.
Our agent interacts with these tools in a multi-turn fashion and leaves a conversation transcript, enhancing interpretability, transparency, and trust---all crucial for scientific research.

\section{Methodology}
\label{sec:methodology}

Our proposed framework, \textbf{\aiscivision}, integrates a Visual Retrieval-Augmented Generation (\visrag) procedure with domain-specific tools, which an LMM uses to classify images in the scientific domain.
In this section, we describe these two components and discuss how \aiscivision uses them during inference.

\subsection{Retrieving Relevant Images with Visual RAG (\visrag)}
To specialize a general-purpose LMM for scientific image classification during inference, we enrich the model's prompt with images relevant to the given test image.
We first encode all available training images into a shared embedding space.
Let $\calD = \set{ (\bfx_i, y_i) }_{i=1}^N$ be a training set of binary labeled images, where $\bfx_i \in \bbR^{H \times W \times C}$ is an image and $y_i \in \set{\pm 1}$ is its label.
We map each image $\bfx_i$ to an embedding $\bfe_i = \phi(\bfx_i) \in \bbR^d$ using an embedding model $\phi : \calX \to \bbR^d$ on the set of images $\calX$.
This embedding model could be a pre-trained image embedding model, for instance, CLIP~\citep{radford2021learning}, that ensures that similar visual content has similar embeddings.
Second, we separate the embeddings into two sets: positive examples $\calE^+ = \set{ \bfe_i \mid y_i = 1 }$ and negative examples $\calE^- = \set{ \bfe_i \mid y_i = -1 }$.
This allows us to enrich the model's prompt with structured context, described below.

On inference, we embed an input test image $\xtest$ to get embedding $\etest = \phi(\xtest)$.
We then retrieve relevant images to enrich the LMM's context by computing the cosine similarity of the test image embedding $\xtest$ with all positive embeddings $\calE^+$, and with all negative embeddings $\calE^-$.
By ranking all embeddings according to the cosine similarity, we obtain the most similar positive example $\bfepos$ and negative example $\bfeneg$ as follows:
\begin{align*}
    \bfepos &\coloneqq \argmax_{\bfe_i \in \calE^+} \cos(\etest, \bfe_i) \quad \text{and} 
    \quad \bfeneg \coloneqq \argmax_{\bfe_i \in \calE^-} \cos(\etest, \bfe_i), \quad \text{where }
    \cos(\etest, \bfe_i) = \frac{\etest \cdot \bfe_i}{ \norm{\etest} \norm{\bfe_i} }.
\end{align*}

We then provide the images $\bfxpos$ and $\bfxneg$ of the respective embeddings $\bfepos$ and $\bfeneg$ to the LMM in its prompt. 
This enables the model to evaluate images it might not have been trained on, in our case, scientific images.
Adding both positive and negative examples provides relevant visual features that characterize the domain-specific classification task, effectively helping to ground the model's reasoning.
Hence, this \visrag approach facilitates more accurate and context-aware inference, leveraging the structure inherent in the classification task.

\subsection{Domain-specific Interactive Tools}
We leverage expert-designed tools for each classification task, empowering the LMM to refine its predictions by interacting with these tools.
These tools mimic transformations and ``expert advice'' that a human would use to manipulate, inspect, and analyze images, before attempting to classify an image.
Therefore, by interacting with these tools in the \aiscivision framework, the LMM acts akin to an interactive AI agent making informed and interpretable decisions.

Generally, we define a tool $T$ as a function on images $\calX$, with outputs as images $\calX$ or a real-valued scalar in $\bbR$.
That is, such a tool either transforms an image $\bfx \in \calX$ to another image $T(\bfx) \in \calX$ or returns a numeric output $T(\bfx) \in \bbR$ such as a confidence score from an external model.
For example, a tool $T_{\text{ML}}$ might use an externally-trained Machine Learning model with parameters $\theta$ to predict the probability of a label given the image: $T_{\text{ML}} = \Pr [y = 1 \mid \bfx; \theta]$.
Other tools $T_{\text{br}}$ or $T_{\text{co}}$ might adjust brightness or contrast by some value $\alpha$ such that $T_{\text{br}} = \text{AdjustBrightness}(\bfx, \alpha)$ or $T_{\text{co}} = \text{IncreaseConstrast}(\bfx, \alpha)$.

For each image classification task, we define a set of tools $\calT = \set{ T_1, \dots, T_K }$ and provide their descriptions in natural language as a prompt to the LMM.
At any turn $i$ in the conversation, the model submits a request for a tool $T_i \in \calT$, which can either transform the image $\bfx' = T_i (\bfx)$ or return a numeric value.
The \aiscivision parses this request and returns the tool's result as a prompt, with the transformed image or the numeric value described in a sentence.
In essence, this is similar to ReAct~\citep{yao2023react} in that we use a hardcoded prompt template for the response.
Iterative use of domain-specific tools enables the agent in our \aiscivision framework to refine predictions in a context-aware manner, not only improving accuracy (see \Cref{sec:experiments}) but also producing a transcript of the agent's reasoning.
Such a transcript provides interpretable insight into our framework's reasoning, which is crucial for applications in scientific discovery.

\subsection{Inference Process in the \aiscivision Framework}

Given an input test image, \aiscivision enriches the model's prompt with \visrag and descriptions of the tools, which the model calls in subsequent turns of the conversation.
Finally, the model outputs a classification label with a probability score indicating its confidence.

We design the initial system prompt to reflect the specified domain (see \Cref{app:example_transcripts} for example transcripts).
First, we use the \visrag approach to retrieve the most similar positive and negative examples from the training set, $\bfxpos$ and $\bfxneg$ respectively.
We then describe the set $\calT$ of available domain-specific tools, and encourage the model to use them to obtain more context during inference.
After a few conversation turns, the LMM responds with a binary prediction and a confidence score.
The transcript of interactions represents a record of justifications at the inference stage, which a domain expert can review after the fact.
In this way, \aiscivision presents an interactive AI agent that uses domain-specific knowledge to not only classify scientific images but also justify its underlying reasoning in natural language.

\begin{figure*}[t]
    \centering
    \includegraphics[width=\textwidth]{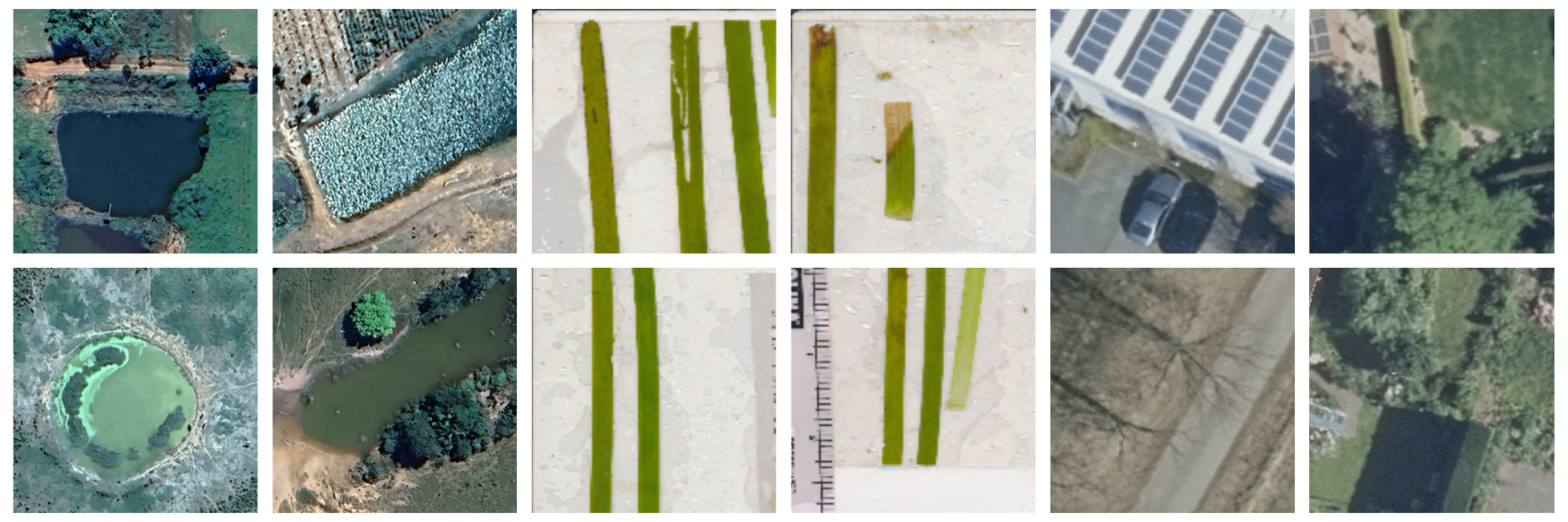}
    \caption{Four examples images from each dataset: \dsaquaculture, \dseelgrass, and \dssolar (left to right).
    The top row are all positive examples (aquaculture ponds present, eelgrass plant diseased, and solar panels present), and the bottom row are all negative examples.
    }
    \label{fig:ds}
\end{figure*}

\section{Experiments}
\label{sec:experiments}

We extensively evaluate our \aiscivision framework on three image datasets from scientific applications: detecting the presence of aquaculture ponds~\citep{greenstreet2023detecting}, diseased eelgrass~\citep{rappazzo2021eelisa}, and solar panels~\citep{lacoste2023geobenchfoundationmodelsearth}.
We compare against natural baselines and conduct ablation studies on components of \aiscivision.
We discuss our experimental results in this section.

\paragraph{Datasets and Experimental Setup}

We provide a brief overview of the three datasets and their significance to environmental research, along with a summary of the domain-specific tools used in the \aiscivision framework (see \Cref{app:dataset_specific_tools} for a full list of tools).

\begin{enumerate} 
    \item \textbf{\dsaquaculture Pond Detection.} Aquaculture, vital for the global food supply, requires careful monitoring from satellite imagery.
    This poses challenges due to the varied appearance of water bodies.
    The \dsaquaculture dataset contains 799 images ($640\times640$) from Rondônia, Brazil~\citep{greenstreet2023detecting}, with $\approx 20\%$ containing aquaculture ponds. 
    Since the metadata includes geospatial location data, in \aiscivision we define tools like zoom and pan, utilizing real-time Google Maps API. 

    \item \textbf{\dseelgrass Wasting Disease Detection.} Eelgrass (\textit{Zostera marina}), essential for coastal ecosystems, faces threats from Eelgrass Wasting Disease (EWD)~\citep{groner2016plant}.
    The \dseelgrass dataset contains 9887 images ($128 \times 128$) from Washington state, with $\approx 45\%$ showing diseased eelgrass~\citep{rappazzo2021eelisa}. 
    We incorporate tools like contrast and sharpening adjustments. 
    
    \item \textbf{\dssolar Panel Detection.} The open-source \dssolar dataset tracks solar panel adoption with satellite images~\citep{lacoste2023geobenchfoundationmodelsearth}.
    It contains 11814 images ($320 \times 320$), with $\approx 15\%$ containing solar panels.
    We use similar image enhancement tools as the \dseelgrass dataset, since geospatial metadata is absent. 
    \end{enumerate}

We test on 100 randomly subsampled examples from each dataset's test set for consistent evaluation across all methods.
To balance the cost constraints of LMM experimentation, we use this small test set for robust experiments and ablation studies.
All methods are evaluated in low-labeled ($20\%$) and full-labeled ($100\%$) data settings, on Accuracy, F1-score, and Area Under Curve (AUC) metrics.

\paragraph{\aiscivision method}
We use the GPT-4o as the framework's LMM, and attach the two components \visrag and domain-specific tools for each dataset.
We provide text descriptions of the available tools as prompts to the LMM, and instruct how to request tool use.
We encourage the LMM to use at least 3 tools during inference, and instruct it to submit a classification decision with a confidence value in 4 conversation turns. All embeddings for VisRAG are computed via CLIP, and the cosine similarity is used to gauge the similarity between images. 
We include example transcripts and our prompts for all datasets in \Cref{app:example_transcripts}.

\begin{table}[!htbp]
    \centering

    \resizebox{\textwidth}{!}{
    \begin{tabular}{l|ccc|ccc|ccc|ccc|ccc|ccc}
        \toprule
        \multirow{3}{*}{} & \multicolumn{6}{c|}{\dsaquaculture} & \multicolumn{6}{c|}{\dseelgrass} & \multicolumn{6}{c}{\dssolar} \\
        \cmidrule{2-19}
        & \multicolumn{3}{c|}{20\%} & \multicolumn{3}{c|}{100\%} & \multicolumn{3}{c|}{20\%} & \multicolumn{3}{c|}{100\%} & \multicolumn{3}{c|}{20\%} & \multicolumn{3}{c}{100\%} \\
        \cmidrule{2-19}
        Methods & Acc & F1 & AUC & Acc & F1 & AUC & Acc & F1 & AUC & Acc & F1 & AUC & Acc & F1 & AUC & Acc & F1 & AUC \\
        \midrule
        $k$-NN & 0.84 & 0.65 & 0.83 & 0.80 & 0.52 & 0.76 & 0.71 & 0.66 & 0.79 & 0.78 & 0.72 & 0.86 & 0.96 & 0.80 & 0.91 & 0.95 & 0.76 & 0.94 \\
        CLIP-ZeroShot & 0.82 & 0.00 & 0.56 & 0.82 & 0.00 & 0.56 & 0.60 & 0.17 & 0.54 & 0.60 & 0.17 & 0.54 & 0.63 & 0.33 & 0.65 & 0.63 & 0.33 & 0.65 \\
        CLIP+MLP & 0.85 & 0.67 & 0.91 & 0.86 & 0.63 & \textbf{0.93} & \textbf{0.82} & 0.72 & 0.88 & 0.80 & 0.74 & 0.92 & \textbf{0.98} & 0.91 & \textbf{1.00} & \textbf{0.99} & \textbf{0.96} & \textbf{1.00} \\
        \aiscivision & \textbf{0.90} & \textbf{0.78} & \textbf{0.95} & \textbf{0.92} & \textbf{0.81} & \textbf{0.93} & 0.81 & \textbf{0.73} & \textbf{0.89} & \textbf{0.84} & \textbf{0.80} & \textbf{0.95} & \textbf{0.98} & \textbf{0.92} & \textbf{1.00} & 0.97 & 0.88 & \textbf{1.00} \\
        \bottomrule
    \end{tabular}
    }
    \caption{Our \aiscivision framework (GPT4o + \visrag + Tools) consistently outperforms zero-shot and supervised methods across all datasets and metrics.
    We evaluate all methods in both low-labeled ($20\%$) and full-labeled ($100\%$) regimes.
    The CLIP+MLP supervised model, trained on the same images, is used as a tool within \aiscivision, providing prediction probabilities upon request by the LMM.
    We observe that the na\"ive $k$-NN baseline outperforms CLIP-ZeroShot and is competitive with CLIP+MLP.
    This highlights the importance of leveraging domain-specific structure in scientific image classification over general-purpose LMMs, justifying the \visrag component in \aiscivision.
    We further find that \aiscivision outperforms CLIP+MLP, which is provided as a tool in our framework, implying that \aiscivision does not solely rely on this tool.
    Precision and recall for all methods are detailed in \Cref{tab:prec_recall}.}
    \label{tab:main}
\end{table}

\paragraph{Baselines}
A key component of \aiscivision is \visrag, which retrieves the most similar positive and negative examples during inference before prompting the LMM.
Hence, a natural baseline is na\"ive $k$-Nearest Neighbor (\textbf{$k$-NN} with $k=3$) using CLIP embeddings.
We also include \textbf{CLIP-ZeroShot} as a baseline: we classify a test image by comparing the cosine similarities of CLIP text embeddings of the two labels with the CLIP image embedding.
As CLIP-ZeroShot does not rely on image features specific to the scientific domain, we evaluate another baseline \textbf{CLIP+MLP} for binary classification, where we train a 2-layer Multi-Layer Perceptron (MLP) on top of frozen CLIP image embeddings.
The MLP has a hidden layer with 256 units and ReLU activation function, and is trained for 10 epochs using the Adam optimizer (learning rate $0.01$ and batch-size 32).
We find that \aiscivision consistently outperforms all baselines on all three datasets, in both low- and full-labeled data regimes (\Cref{tab:main}).
\footnote{In \Cref{tab:main}, note that CLIP-ZeroShot does not use any training images from the datasets, hence the same values of metrics between the low- and full-labeled data regimes.
In our experiments, CLIP-ZeroShot does indeed attain $0.0$ F1 score for the \dsaquaculture dataset, because it does not predict any true positives (aquaculture ponds present).
It has been shown that CLIP embeddings perform poorly zero-shot on satellite images~\citep{radford2021learning,mall2024remote}.
}

\paragraph{Ablation studies and tool efficacy}
We conduct ablation studies on the two components of \aiscivision: \visrag and domain-specific tools.
We compare the following variants with \aiscivision (GPT4o + \visrag + Tools): (1) \textbf{GPT4o-ZeroShot} that predicts a label and confidence score after the initial prompt\footnote{GPT4o generations are \href{https://platform.openai.com/docs/advanced-usage/reproducible-outputs}{inherently random} even after setting a seed, system fingerprint, the temperature to 0. This leads to slight variability in the values of metrics.}, (2) \textbf{GPT4o + \visrag} that uses only \visrag to retrieve examples relevant to the test image, (3) \textbf{GPT4o + Tools} that uses only our domain-specific tools.
We report these ablation experiment results in \Cref{tab:ablations}.
These ablation experiments allow us to isolate and evaluate the effects of retrieval through \visrag and usage of domain-specific tools.

\begin{figure}[!b]
    \centering
    \includegraphics[width=\textwidth]{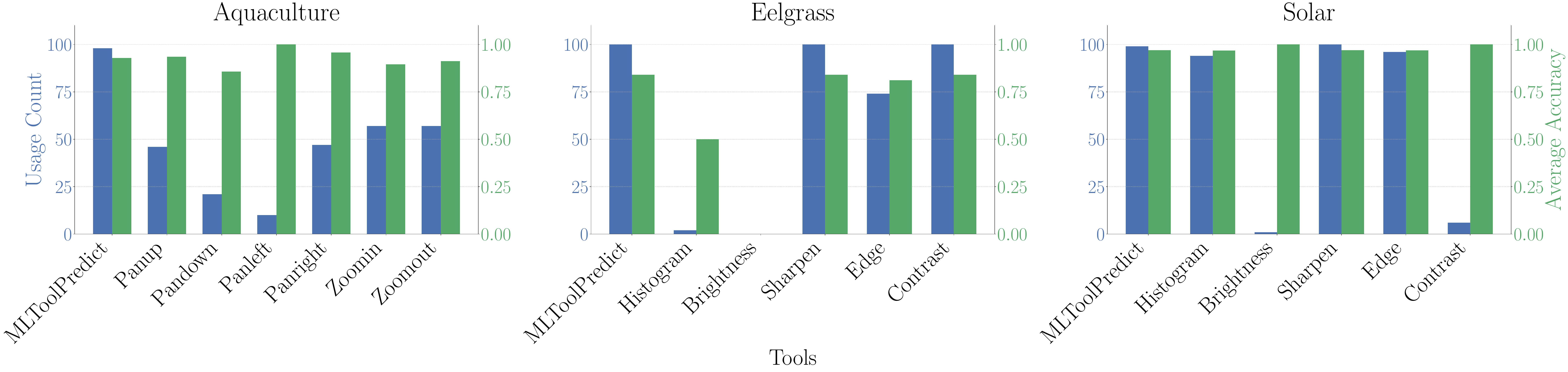}
    \caption{
    We analyze the frequency of tool use in \aiscivision for each dataset, and their effects on the final classification result.
    {\bf\color{blue!50!black}Blue} bars indicate the number of tool calls during inference on our test sets of 100 images, whereas {\bf\color{green!50!black}green} bars indicate the mean classification accuracy \textit{when} the specified tool was called in a conversation in \textit{any} round.
    Note that the LMM can call a tool multiple times in the same conversation.
    Across all datasets, the LMM almost always requests the supervised model tool ``MLToolPredict'' but does not solely rely on the returned results.
    The LMM heavily relies on geospatial tools for the \dsaquaculture dataset, since these tools return additional information from the vicinity.
    It is curious that the LMM does not use the AdjustBrightness tool at all and uses HistogramEqualization sparingly for the other two datasets.
    All tools are described in \Cref{app:dataset_specific_tools}.}
    \label{fig:tool_usage}
\end{figure}

Moreover, we log all inference transcripts and frequencies of tool usage in \aiscivision.
We then analyze the effect of different tools on the agent's classification accuracy, reported in \Cref{fig:tool_usage}, offering insight into their efficacy for the agent's decision-making process.

\begin{table}[!t]
    \centering
    \resizebox{\textwidth}{!}{
     \begin{tabular}{l|ccc|ccc|ccc|ccc|ccc|ccc}
        \toprule
        \multirow{3}{*}{} & \multicolumn{6}{c|}{\dsaquaculture} & \multicolumn{6}{c|}{\dseelgrass} & \multicolumn{6}{c}{\dssolar} \\
        \cmidrule{2-19}
        & \multicolumn{3}{c|}{20\%} & \multicolumn{3}{c|}{100\%} & \multicolumn{3}{c|}{20\%} & \multicolumn{3}{c|}{100\%} & \multicolumn{3}{c|}{20\%} & \multicolumn{3}{c}{100\%} \\
        \cmidrule{2-19}
        Methods & Acc & F1 & AUC & Acc & F1 & AUC & Acc & F1 & AUC & Acc & F1 & AUC & Acc & F1 & AUC & Acc & F1 & AUC \\
        \midrule
        GPT4o-ZeroShot & 0.86 & 0.67 & 0.85 & 0.85 & 0.62 & 0.86 & 0.77 & \textbf{0.74} & 0.86 & 0.74 & 0.71 & 0.83 & 0.93 & 0.77 & 0.99 & 0.91 & 0.73 & 0.99 \\
        GPT4o + \visrag & 0.87 & 0.68 & 0.89 & 0.85 & 0.67 & 0.90 & 0.77 & \textbf{0.74} & 0.86 & \textbf{0.86} & \textbf{0.83} & 0.92 & 0.97 & 0.89 & 0.99 & \textbf{0.97} & \textbf{0.89} & 0.99 \\
        GPT4o + Tools & 0.88 & 0.74 & 0.88 & 0.90 & 0.76 & 0.91 & 0.79 & \textbf{0.74} & \textbf{0.90} & 0.77 & 0.74 & 0.90 & 0.96 & 0.86 & \textbf{1.00} & 0.96 & 0.86 & \textbf{1.00} \\
        \aiscivision & \textbf{0.90} & \textbf{0.78} & \textbf{0.95} & \textbf{0.92} & \textbf{0.81} & \textbf{0.93} & \textbf{0.81} & 0.73 & 0.89 & 0.84 & 0.80 & \textbf{0.95} & \textbf{0.98} & \textbf{0.92} & \textbf{1.00} & \textbf{0.97} & 0.88 & \textbf{1.00} \\
        \bottomrule
    \end{tabular}
    }
    \caption{
    We conduct ablation studies on each component of our framework, \aiscivision (GPT-4o + \visrag + Tools), and find that it generally outperforms the ablation methods.
    Even so, the isolated benefits of \visrag and Tools are notable in \dseelgrass and \dssolar datasets.
    While geospatial tools are particularly effective in the \dsaquaculture dataset of satellite images, the benefits of \visrag complements those of domain-specific tools to significantly outperform either component separately.
    On a closer look at inference transcripts of the \dseelgrass and \dssolar datasets, we observe that the supervised model tool can sometimes introduce a bias that degrades \aiscivision's performance (see transcripts in \Cref{app:abl_failure}).
    We leave it to future work to improve upon our image enhancement tools and develop techniques to select the optimal toolset.
    }
    \label{tab:ablations}
\end{table}

\section{Deployed Application}
\label{sec:deployed_application}

\begin{figure}[!t]
    \centering
    \includegraphics[width=.6\textwidth]{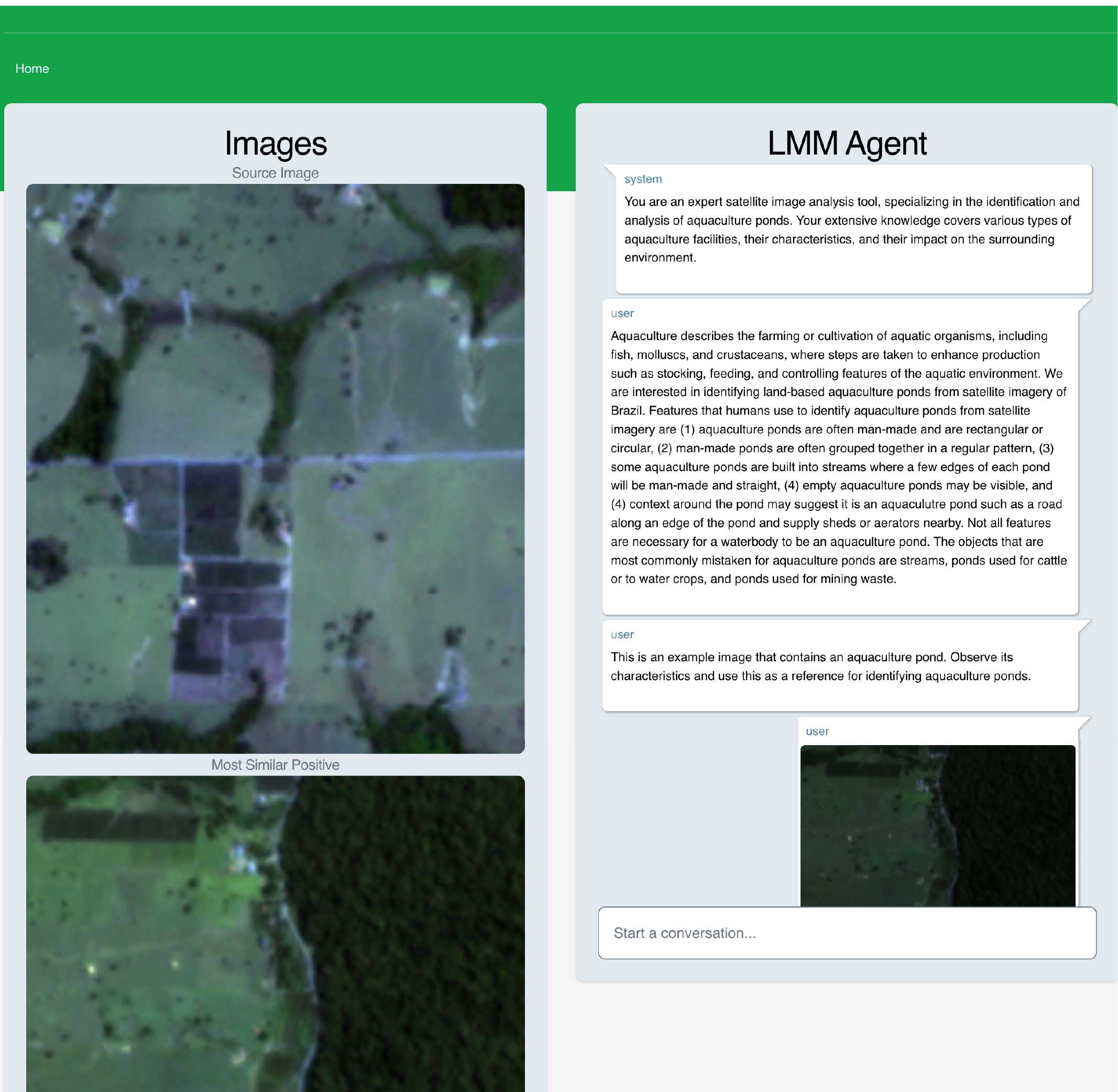}
    \caption{Screenshot of the \aiscivision web application for aquaculture pond detection.
    The application allows users to upload satellite images, interact with visual tools such as zoom and contrast adjustment, and view predictions alongside retrieved similar examples. Additionally, the produced transcript is conversable in a `Chat-GPT' like manner. Allowing the user to ask questions about the inference steps, or to even give rich natural language feedback about errors the model might have made. For future work, we aim to incorporate this feedback into the VisRAG, enabling an ever-learning RAG system that continually improves our model.}
    \label{fig:web_app}
\end{figure}

\aiscivision is not just a conceptual framework, we have deployed it as a fully functional web application to detect aquaculture ponds, enabling real-time and scalable use by ecologists (see a screenshot of the interface in \Cref{fig:web_app}).
When ecologists upload a test image to the web application, the \aiscivision framework prompts the LMM to detect if the image contains aquaculture ponds.
Importantly, \aiscivision provides the detailed transcript that outlines the LMM's reasoning and tool use.
\aiscivision's deployed interface not only provides an accurate classification result but also a transparent decision-making process---crucial for expert ecologists to validate the framework.

This practical prototype opens the potential for expert ecologists to return feedback on both the classification results and the underlying reasoning.
We aim to integrate this feedback algorithmically to improve the \visrag component of \aiscivision, building a framework that improves with expert use.
There are many exciting research directions: seamlessly incorporating feedback loops, utilizing enhanced tools and understanding their interactions, and improving the framework's adaptability to other scientific applications.

\section{Discussion}

Our \aiscivision framework combines robust prediction capabilities, transparency, and adaptability, offering a practical approach for AI use in scientific contexts. 
By delivering accurate and context-aware predictions along with a full reasoning transcript, \aiscivision serves as a valuable research partner across different applications.
Transcripts enhance the accountability and traceability of model outputs, enabling researchers to validate decision-making processes.
This is particularly important in scientific research, for instance in conservation efforts in complex ecosystems like the Amazon Basin, where accurate spatial data is crucial for sustainability.
The combination of predictions and interpretable transcripts can help advance scientific discovery, support regulatory compliance by providing a clear record of inferences, and aid education by giving concrete examples of classification processes.

\paragraph{Limitations and Future Work}
While \aiscivision offers significant benefits in providing transparent reasoning and the potential for enhanced accuracy, it comes with a trade-off: using off-the-shelf LMMs for inference is financially expensive, compared to traditional machine learning methods.
For our future work, we aim to actively develop our web application to  continue to collect expert feedback on the LMM agent's reasoning through a ChatGPT-style interface. Experts can provide real-time feedback and corrections, which will be stored within the \visrag component to improve the LMM agent's performance with use. We envision the system continuing to learn as experts interact with the agent, and provide rich natural language feedback.  Beyond refining our approach for image data, we also plan to test and extend our method to other modalities, such as sound or any tokenizable input that can be incorporated into an LMM.

\section*{Acknowledgements}
This project is partially supported by the National Science Foundation (NSF); the Eric and Wendy Schmidt AI in Science Postdoctoral Fellowship, a program of Schmidt Sciences, LLC; the National Institute of Food and Agriculture (USDA/NIFA); the Air Force Office of Scientific Research) (AFOSR), and Toyota Research Institute (TRI).

\bibliography{refs}
\bibliographystyle{arxiv}

\newpage

\appendix

\section{Domain-specific Tools}
\label{app:dataset_specific_tools}

The tools developed for each dataset are designed to mimic the strategies that human experts would use when performing these specific classification tasks.
For the Eelgrass wasting disease and Solar panel detection tasks, the tools focus on image enhancement techniques that a human might naturally apply when examining an image closely.
These include adjusting brightness, contrast, and sharpness, as well as applying edge detection and histogram equalization.
Such manipulations can reveal features or patterns that might be helpful for accurate classification, much like how a human expert might squint, tilt their head, or adjust lighting to better perceive important details.

In contrast, the aquaculture pond detection task presents a unique challenge where the presence of water bodies isn't always immediately apparent from a single satellite image.
To address this, the tools for this task are designed to emulate how a human would interact with digital maps.
They include options to pan and zoom, allowing for exploration of the surrounding area and changes in perspective.
This approach mirrors how a human expert might navigate digital mapping software, zooming out to get context from the broader landscape, then zooming in on areas of interest to confirm the presence of aquaculture ponds.
By providing these map navigation tools, the framework can gather additional spatial information that may be crucial for accurate classification, especially in cases where a single image view might be ambiguous or inconclusive.

We detail the full list of tools for each dataset in our \aiscivision framework.

\subsection{Tools for \dsaquaculture dataset}

\begin{itemize}
    \item \textbf{PredictAquaculturePondTool}: Predicts the probability of an aquaculture pond being present in the image using a machine learning model. This tool is particularly helpful when you need a quantitative assessment of the likelihood of aquaculture pond presence in the satellite image.
    
    \item \textbf{PanUpToolRelative}: Pans the view upwards relative to the last image seen.
    
    \item \textbf{PanUpToolAbsolute}: Pans the view upwards relative to the original starting image.
    
    \item \textbf{PanDownToolRelative}: Pans the view downwards relative to the last image seen.
    
    \item \textbf{PanDownToolAbsolute}: Pans the view downwards relative to the original starting image.
    
    \item \textbf{PanLeftToolRelative}: Pans the view to the left relative to the last image seen.
    
    \item \textbf{PanLeftToolAbsolute}: Pans the view to the left relative to the original starting image.
    
    \item \textbf{PanRightToolRelative}: Pans the view to the right relative to the last image seen.
    
    \item \textbf{PanRightToolAbsolute}: Pans the view to the right relative to the original starting image.
    
    \item \textbf{ZoomInToolRelative}: Zooms in on the center of the current view relative to the last image seen.
    
    \item \textbf{ZoomInToolAbsolute}: Zooms in on the center of the original view relative to the original starting image.
    
    \item \textbf{ZoomOutToolRelative}: Zooms out from the current view relative to the last image seen.
    
    \item \textbf{ZoomOutToolAbsolute}: Zooms out from the original view relative to the original starting image.
\end{itemize}

\subsection{Tools for \dseelgrass dataset}

\begin{itemize}
    \item \textbf{AdjustBrightnessTool}: Adjusts the brightness of the image by 50\%. This tool can help when the image is too dark or too bright, allowing for better visibility of disease symptoms on the eelgrass blade.
    
    \item \textbf{SharpenTool}: Sharpens the image to enhance edges and details. This tool is useful for making subtle features more prominent, which can help in identifying signs of eelgrass wasting disease.
    
    \item \textbf{EdgeDetectionTool}: Applies edge detection to the image, highlighting boundaries and features. This can help in identifying lesions or patterns associated with eelgrass wasting disease.
    
    \item \textbf{IncreaseContrastTool}: Increases the contrast of the image by 50\%. This tool can be helpful when the image appears too flat or when you need to enhance the visibility of subtle details, especially in cases where eelgrass wasting disease symptoms might be hard to distinguish.
    
    \item \textbf{DecreaseContrastTool}: Decreases the contrast of the image by 50\%. This tool can be useful when the image appears too harsh or when you want to reduce the intensity of bright areas, which might help in identifying overall patterns or structures in the eelgrass.
    
    \item \textbf{PredictEelgrassWastingDiseaseTool}: Predicts the probability of eelgrass wasting disease in the image using a machine learning model. This tool is particularly helpful when you need a quantitative assessment of the likelihood of disease presence in the eelgrass sample.
    
    \item \textbf{HistogramEqualizationTool}: Enhances the contrast of the image using histogram equalization. This can help in making features more distinguishable, which is beneficial for detecting eelgrass wasting disease symptoms.
\end{itemize}

\subsection{Tools for \dssolar dataset}

\begin{itemize}
    \item \textbf{HistogramEqualizationTool}: Enhances the contrast of the image using histogram equalization. This can help in making features more distinguishable, which is beneficial for detecting solar panels and potential defects.
    
    \item \textbf{AdjustBrightnessTool}: Adjusts the brightness of the image by 50\%. This tool can help when the image is too dark or too bright, allowing for better visibility of solar panels and their features.
    
    \item \textbf{SharpenTool}: Sharpens the image to enhance edges and details. This tool is useful for making subtle features more prominent, which can help in identifying solar panels and potential defects.
    
    \item \textbf{EdgeDetectionTool}: Applies edge detection to the image, highlighting boundaries and features. This can help in identifying the outlines of solar panels and potential defects or anomalies.
    
    \item \textbf{IncreaseContrastTool}: Increases the contrast of the image by 50\%. This tool can be helpful when the image appears too flat or when you need to enhance the visibility of subtle details, especially in cases where solar panels might be hard to distinguish from their surroundings.
    
    \item \textbf{DecreaseContrastTool}: Decreases the contrast of the image by 50\%. This tool can be useful when the image appears too harsh or when you want to reduce the intensity of bright areas, which might help in identifying overall patterns or structures in the solar panel array.
    
    \item \textbf{PredictSolarPanelTool}: Predicts the probability of a solar panel being in the image using a machine learning model. This tool is particularly helpful when you need a quantitative assessment of the likelihood of a solar panel being present in the image.
\end{itemize}

\newpage 

\section{Precision and Recall Results}
\label{app:prec_rec}

In addition to the metrics reported in the main paper in \Cref{tab:main}, we report the Precision and Recall metrics in \Cref{tab:prec_recall}.

\begin{table}[!h]
    \centering

    \resizebox{\textwidth}{!}{
        \begin{tabular}{l|cc|cc|cc|cc|cc|cc}
        \toprule
        \multirow{3}{*}{} & \multicolumn{4}{c|}{\dsaquaculture} & \multicolumn{4}{c|}{\dseelgrass} & \multicolumn{4}{c}{\dssolar} \\
        \cmidrule{2-13}
        & \multicolumn{2}{c|}{20\%} & \multicolumn{2}{c|}{100\%} & \multicolumn{2}{c|}{20\%} & \multicolumn{2}{c|}{100\%} & \multicolumn{2}{c|}{20\%} & \multicolumn{2}{c}{100\%} \\
        \cmidrule{2-13}
        Methods & Prec & Rec & Prec & Rec & Prec & Rec & Prec & Rec & Prec & Rec & Prec & Rec \\
        \midrule
        $k$-NN & 0.68 & 0.62 & 0.61 & 0.46 & 0.58 & 0.76 & 0.67 & 0.78 & \textbf{1.00} & 0.67 & 0.89 & 0.67 \\
        CLIP-ZeroShot & 0.00 & 0.00 & 0.00 & 0.00 & 0.57 & 0.10 & 0.57 & 0.10 & 0.25 & 0.47 & 0.25 & 0.47 \\
        CLIP+MLP & 0.71 & 0.62 & 0.86 & 0.50 & \textbf{0.85} & 0.62 & 0.71 & 0.78 & \textbf{1.00} & 0.83 & \textbf{1.00} & 0.92 \\
        GPT4o-ZeroShot & 0.78 & 0.58 & 0.80 & 0.50 & 0.63 & \textbf{0.89} & 0.60 & 0.86 & 0.63 & \textbf{1.00} & 0.57 & \textbf{1.00} \\
        GPT4o + \visrag & \textbf{0.82} & 0.58 & 0.71 & 0.62 & 0.63 & \textbf{0.89} & \textbf{0.76} & \textbf{0.92} & 0.80 & \textbf{1.00} & 0.80 & \textbf{1.00} \\
        GPT4o + Tools & 0.77 & 0.71 & 0.89 & 0.67 & 0.68 & 0.81 & 0.64 & 0.86 & 0.75 & \textbf{1.00} & 0.75 & \textbf{1.00} \\
        \aiscivision & \textbf{0.82} & \textbf{0.75} & \textbf{0.94} & \textbf{0.71} & 0.76 & 0.70 & 0.73 & 0.89 & 0.92 & 0.92 & 0.85 & 0.92 \\
        \bottomrule
    \end{tabular}
    }
    \caption{
    We report values of precision and recall when all methods are tested in both low- and full-labeled training data regimes, that is $20\%$ and $100\%$ respectively.
    Recall that our framework \aiscivision is GPT4o + \visrag + Tools.
    CLIP-ZeroShot does not obtain any true positives for the aquaculture dataset, i.e. it only predicts that no pond is present, resulting in 0.0 values of precision, recall, and F1-score.
    }
    \label{tab:prec_recall}
\end{table}

\section{Example Transcripts}
\label{app:example_transcripts}

In this section we provide the full transcript for one sample inference from each of the datasets.
These contain the full prompts and inference steps for the \aiscivision framework.

\includepdf[scale=0.75,pages=1,pagecommand=\subsection{\dsaquaculture}]{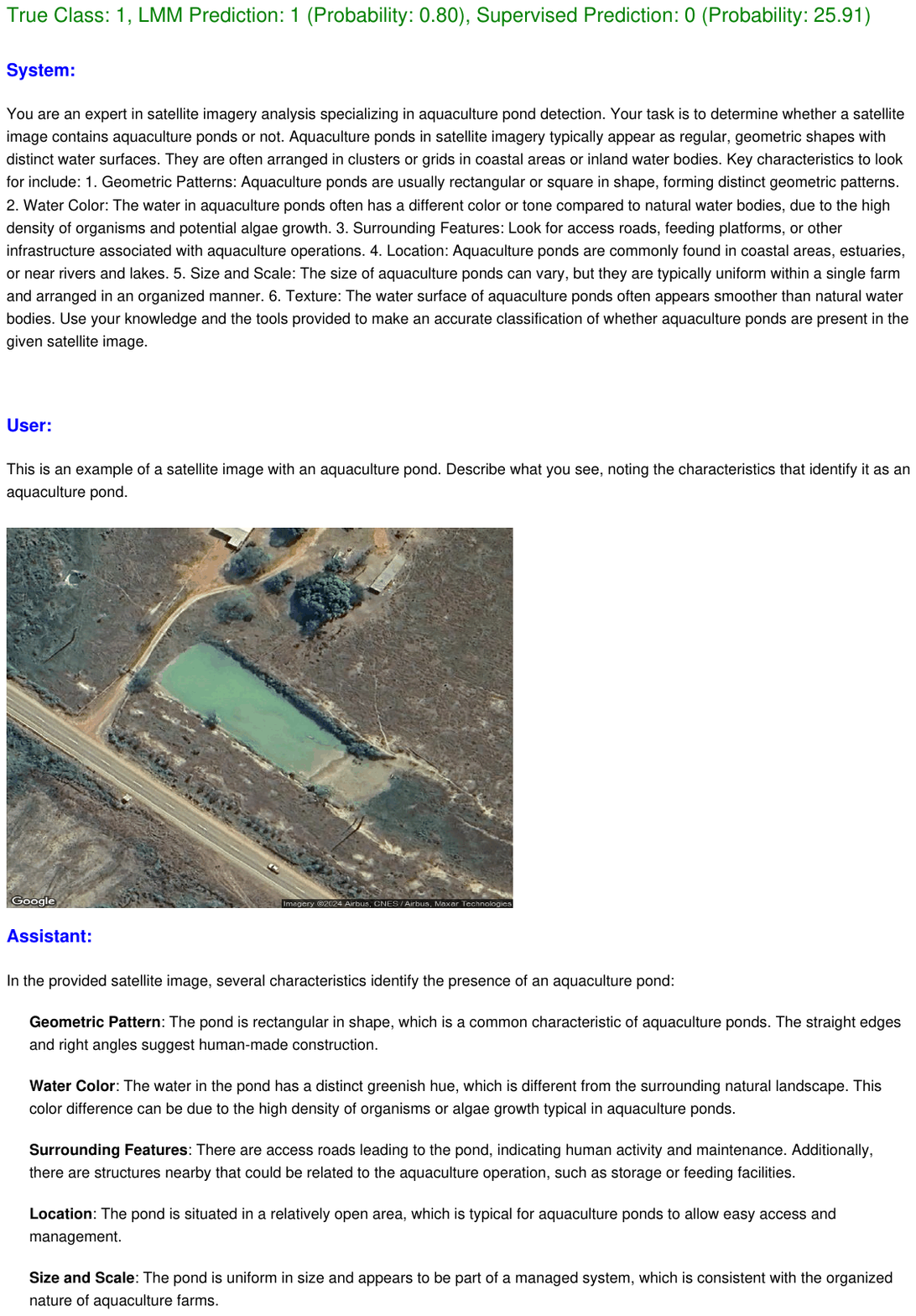}
\includepdf[scale=0.75,pages=2-,pagecommand={}]{appendix/aqua_example.pdf}

\includepdf[scale=0.75,pages=1,pagecommand=\subsection{\dseelgrass}]{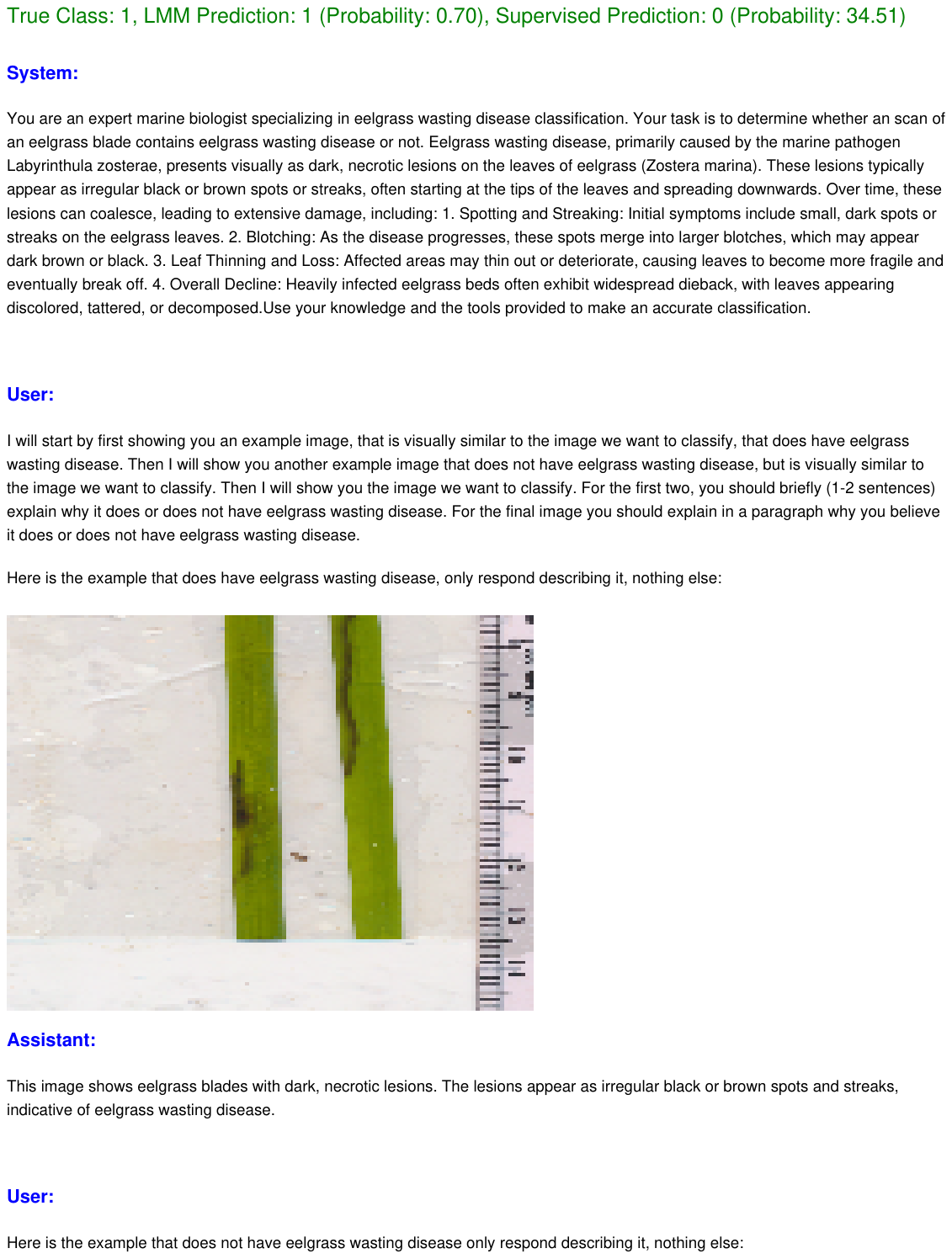}
\includepdf[scale=0.75,pages=2-,pagecommand={}]{appendix/eelgrass.pdf}

\includepdf[scale=0.75,pages=1,pagecommand=\subsection{\dssolar}]{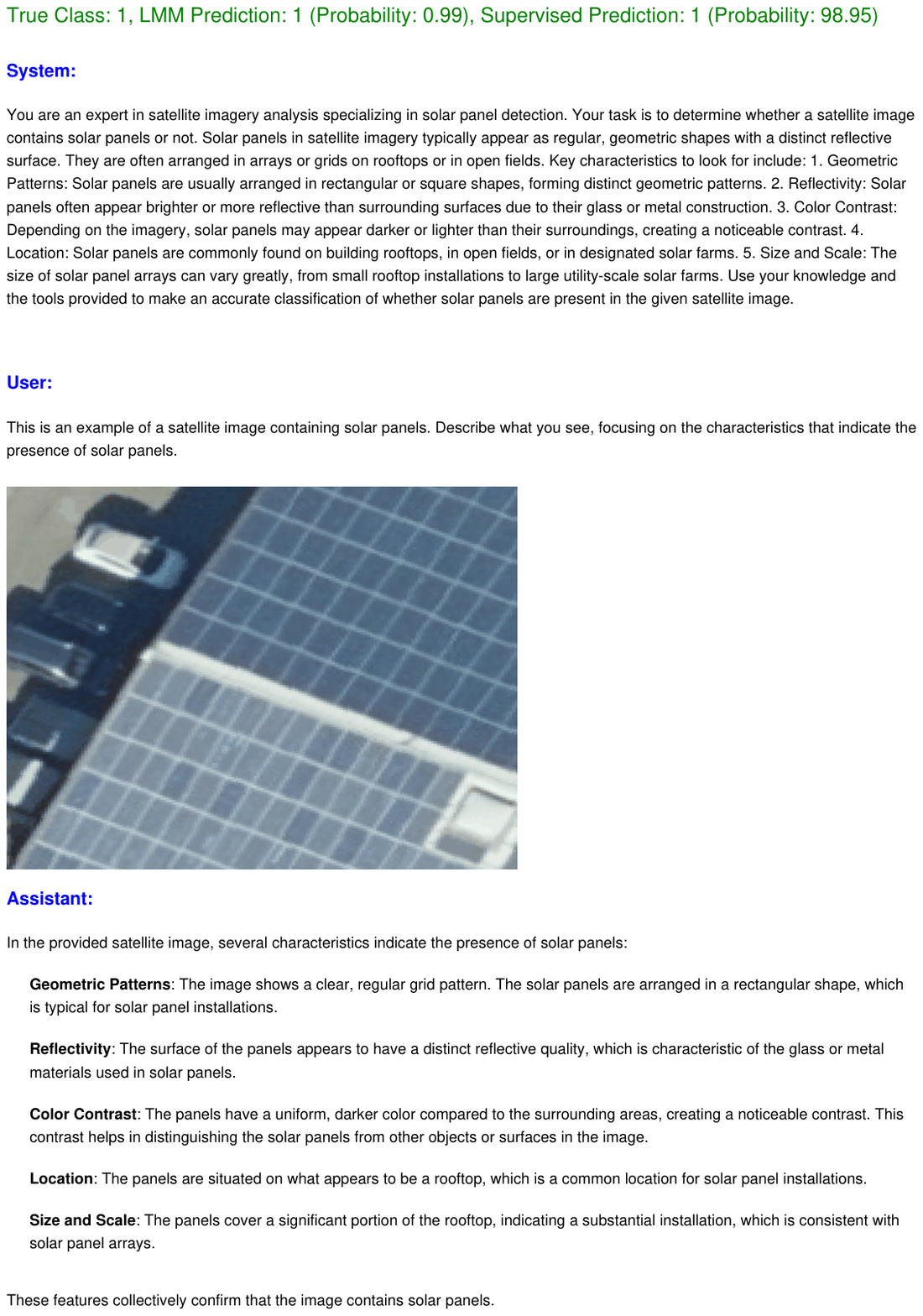}
\includepdf[scale=0.75,pages=2-,pagecommand={}]{appendix/solar_example.pdf}

\section{Ablation Failure Case}
\label{app:abl_failure}

In our analysis of the \dseelgrass dataset, we noticed that the GPT4o + \visrag (without Tools) ablation experiment performed particularly well.
On investigating the inference transcripts more closely, we found an interesting failure case.
After the initial \visrag component, the LMM often predicted the correct class.
The LMM would then request the supervised model tool, and switch its prediction if the supervised model predicted incorrectly.
This ultimately resulted in an incorrect classification.
In these instances, the supervised model tool actually introduced bias in the \aiscivision framework and worsen the performance.

We include transcripts of such an example below.
The GPT4o + \visrag method correctly identifies the diseased eelgrass plant.
But both the GPT4o + Tools and \aiscivision (GPT4o + \visrag + Tools) methods classify incorrectly, and the latter does so by changing its prediction after invoking the supervised model tool.
There are three inference transcripts in this section.

\includepdf[scale=0.75,pages=1,pagecommand=\subsection{GPT4o + \visrag (Correct Prediction)}]{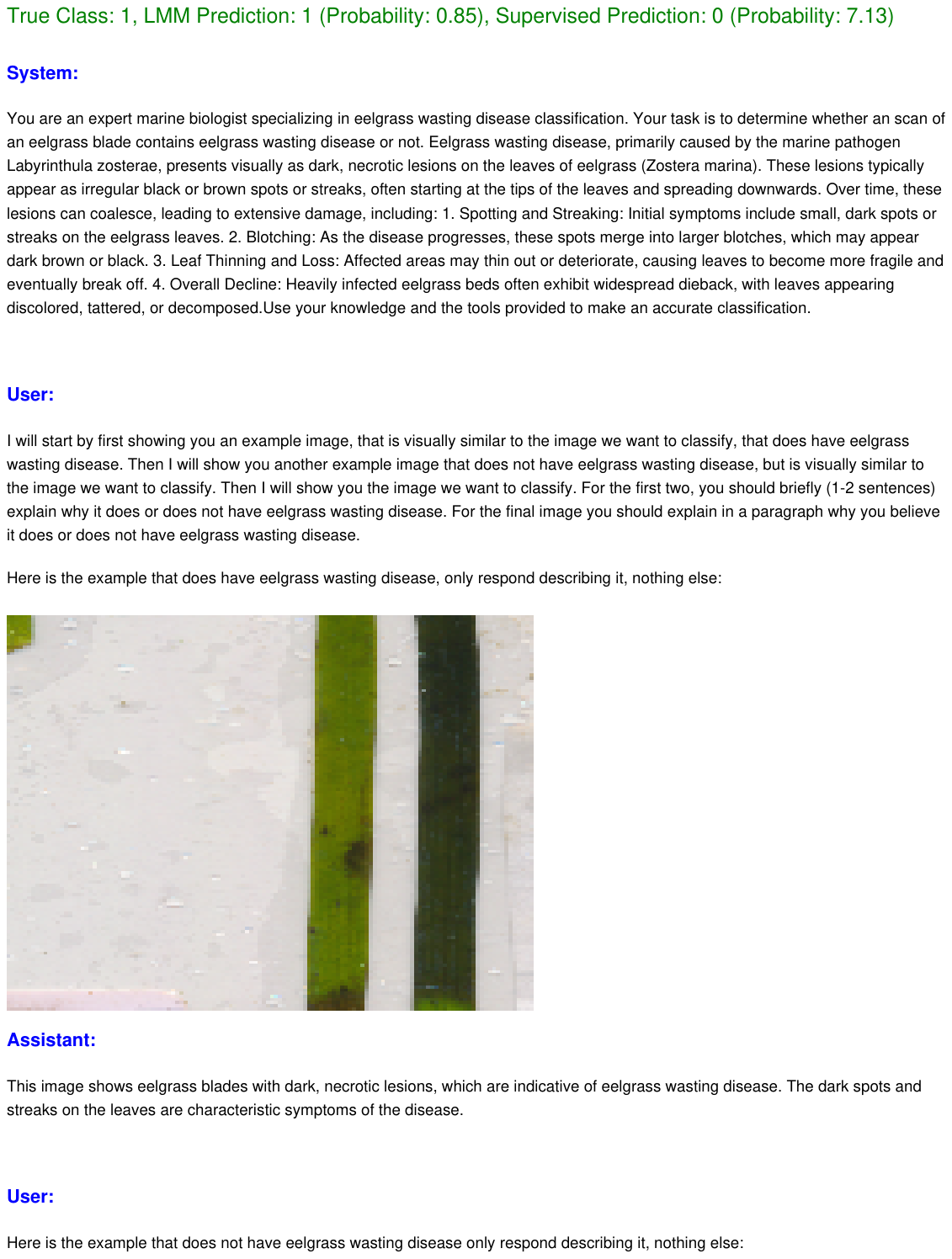}
\includepdf[scale=0.75,pages=2-,pagecommand={}]{appendix/visrag_true1_lmm1_sup0.pdf}

\includepdf[scale=0.75,pages=1,pagecommand=\subsection{GPT4o + Tools (Incorrect Prediction)}]{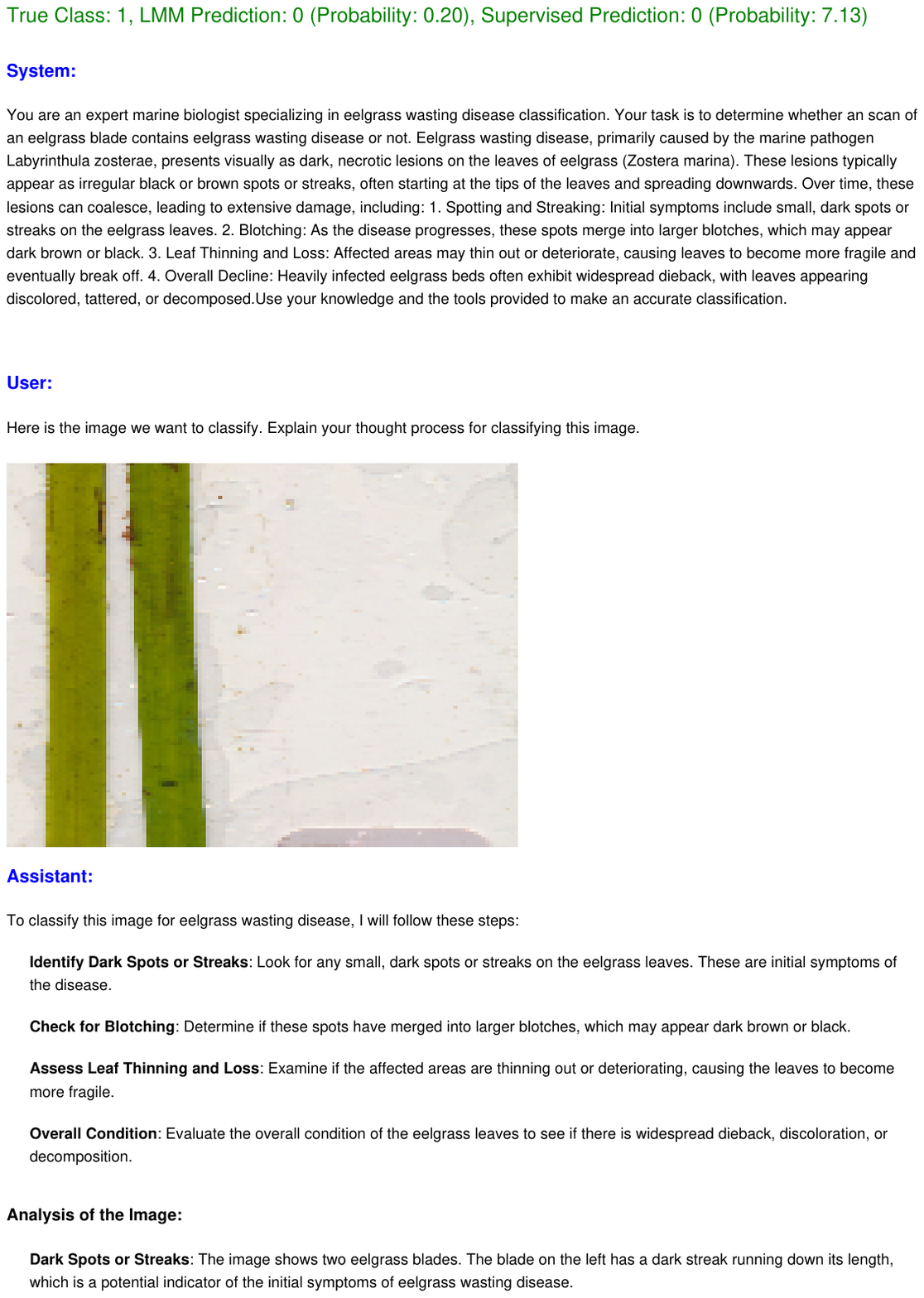}
\includepdf[scale=0.75,pages=2-,pagecommand={}]{appendix/tools_true1_lmm0_sup0.pdf}

\includepdf[scale=0.75,pages=1,pagecommand=\subsection{\aiscivision (GPT4o + \visrag + Tools) (Incorrect Prediction)}]{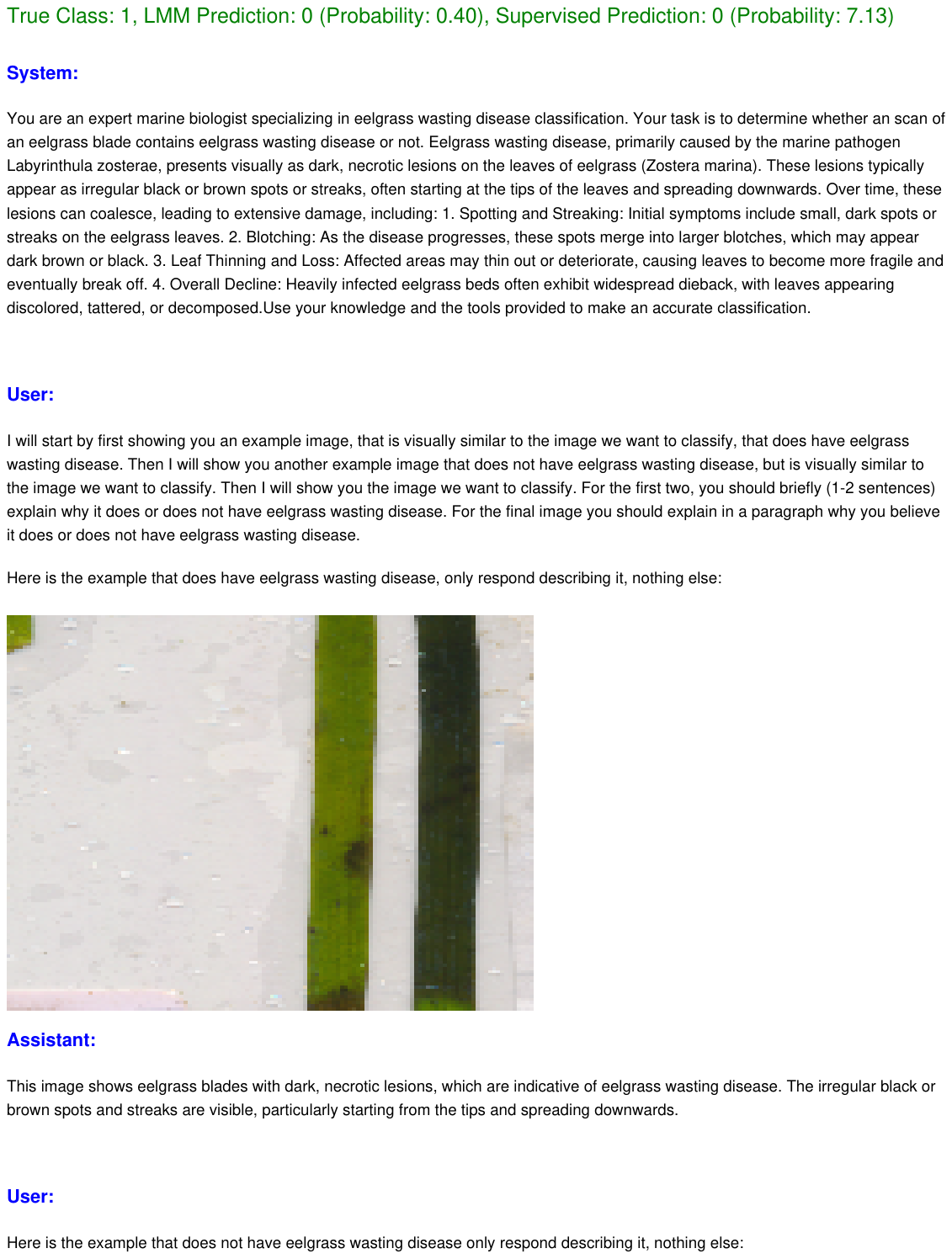}
\includepdf[scale=0.75,pages=2-,pagecommand={}]{appendix/visrag_tools_true1_lmm0_sup0.pdf}

\end{document}